\pdfoutput=1
\documentclass[11pt]{article}

\usepackage[]{latex/acl}

\usepackage{amsmath}
\usepackage{graphicx}
\usepackage{epstopdf} 
\usepackage{multirow}
\usepackage{times}
\usepackage{latexsym}
\usepackage{amssymb}
\usepackage{subfigure}
\usepackage{graphicx}
\usepackage[T1]{fontenc}
\usepackage{hyperref} 
\usepackage{bm}
\usepackage[utf8]{inputenc}
\usepackage{hyperref}
\usepackage{booktabs} 

\usepackage{microtype}
\usepackage[lined,boxed,commentsnumbered]{algorithm2e}
\usepackage{inconsolata}
\title{Benchingmaking Large Langage Models in Biomedical Triple Extraction}

\author{Mingchen Li\textsuperscript{\normalfont  }, Huixue Zhou\textsuperscript{\normalfont  },  \textbf{Rui Zhang}\textsuperscript{\normalfont  }\textbf{}  \\
        \textsuperscript{ }University of Minnesota Twin Cities\\ \textsuperscript{}\{li003378, zhou1742, zhan1386\}@umn.edu\\
        }

\begin{document}
\maketitle

\begin{abstract}
Biomedical triple extraction systems aim to automatically extract biomedical entities and relations between entities. The exploration of applying large language models (LLM) to triple extraction is still relatively unexplored. In this work, we mainly focus on sentence-level biomedical triple extraction.
Furthermore, the absence of a high-quality biomedical triple extraction dataset impedes the progress in developing robust triple extraction systems. To address these challenges, initially, we compare the performance of various large language models.
Additionally, we present \textsc{GIT}, an expert-annotated biomedical triple extraction dataset that covers a wider range of relation types. 




\end{abstract}

\section{Introduction}
Biomedical triple extraction aims to accurately identify relational triples \textit{(head entity, relation, tail entity)} in biomedical text. This process involves recognizing biomedical entity pairs (\textit{head} and \textit{tail}) and extracting the \textit{relation} between these paired entities. It is an important task in natural language processing, as the extracted information can be used to construct knowledge graphs~\cite{fellbaum2010wordnet,li2020multi} and for downstream applications, including link prediction~\cite{li2022hierarchical},  drug repurposing~\cite{zhang2021drug},  question answering~\cite{li2022semantic}.

Most efforts on triple extraction systems focus on employing either the table-filling method~\cite{tang2022unirel,shang2022onerel} or generation methods~\cite{lu2022unified,gao2023easy}. In particular, UniRel~\cite{tang2022unirel} employs the output probability matrix of an auto-encoding language model to assess both entity-entity interactions and entity-relation interactions, while UIE~\cite{lu2022unified}  transfers the different information extraction tasks such as named entity recognition~\cite{li2023far,li2023w}, relation extraction~\cite{sui2023joint,chen2023dialogue} to a unified generation framework. However, the exploration of applying large language models (LLM) to triple
extraction is still relatively unexplored.

Another significant challenge in this field is the scarcity of high-quality datasets.
Current datasets do not offer enough variety in relation types.
For instance, the BioRelEx dataset~\cite{khachatrian2019biorelex} is restricted to just one relation type (Binding),  while the ADE dataset~\cite{gurulingappa2012development} includes two relation types (Drug-adverse effect, Drug-dosage). As indicated by our medical expert, this limited range fails to accurately reflect the complex relationships that exist between biomedical entities. 

So in this work, we system compare the LLM performance on biomedical triple extraction.
Additionally, for the second challenge, we developed a high-quality biomedical triple extraction dataset GIT (\textbf{G}eneral BioMedical and Complementary and \textbf{I}ntegrative Health \textbf{T}riples),
\footnote{Complementary and Integrative Health refers to non-drug therapies} 
characterized by its high-quality annotations and comprehensive coverage of relation types.

\begin{itemize}
     \item  We conduct a thorough analysis of several LLM's performance on three datasets.
     \item  We introduce a biomedical triple extraction dataset GIT with high-quality annotations and comprehensive coverage of relation types.
    
\end{itemize}

\section{Related Work}

\subsection{Triple Extraction Methods}
Most previous studies on triple extraction are directed toward employing either the table-filling method~\cite{tang2022unirel,shang2022onerel,liu2023rexuie} or generation methods~\cite{lu2022unified,gao2023easy,fei2022lasuie, lou2023universal,tan2022query}.  For example,  UniRel~\cite{tang2022unirel}  models entity-entity interactions and
entity-relation interactions in one single interaction map, which is predicted by the output probability matrix of auto-encoding language models, such as BERT~\cite{devlin2018bert}.  OneRel~\cite{shang2022onerel} introduces "BIE" (Begin, Inside, End) signs to indicate the position information of a token within entities and utilizes the labeled relationship in BIE to denote the relationship.
UIE~\cite{lu2022unified} explores the ability to universally model various information extraction tasks and adaptively generate the content of the different tasks.
E2H~\cite{gao2023easy} employs a three-stage approach to enhance entity recognition, relation recognition, and triple extraction capabilities.  Compared with all these studies, our focus is on utilizing the retrieval-augmented language model to prompt the model to accurately retrieve chunk information and generate triples.

\subsection{Biomedical Triple Extraction Datasets}   

There is some prior research~\cite{khachatrian2019biorelex,gurulingappa2012development,cheng2008polysearch,taboureau2010chemprot} that focuses on biomedical triple extraction dataset construction through expert annotation.  For example,
the BioRelEx~\cite{khachatrian2019biorelex} annotates 2,010 sentences extracted from biomedical literature, specifically addressing binding interactions involving proteins and/or biomolecules.  ADE~\cite{gurulingappa2012development} involves the manual annotation of 4,272 sentences from medical reports, specifically focusing on descriptions of drug-related adverse effects.   
DDI~\cite{segura2013semeval} consists of five relation types (mechanism, effect, advice, int, None) extracted from MedLine abstracts.
Despite their success, these datasets lack sufficient variety in relation types to adequately represent the connections among biomedical entities. Therefore, we introduce GM-CIHT in our work to address this limitation. 

\section{ GIT}
Due to the annotation costs, dataset availability is a primary bottleneck for biomedical triple extraction.  Annotating over a thousand pieces of data is a demanding task for experts. In order to reduce the burden of labeling, we propose repurposing an established expert-annotated dataset Medical Triple Classification Dataset (MTCD) \cite{zhang2021drug} from the broader biomedical domain. Simultaneously, we have also undertaken the annotation of 2,450 sentences on non-drug therapies, a task completed by two medical experts.  For the required domain expertise, we enlisted the services of two medical experts who worked for 12 weeks. Each expert worked 25 hours per week at a rate of \$26 per hour.

\subsection{Repurposing MTCD}

MTCD, designed for medical triple classification, consists of 4,352 positive and 2,140 negative pairs sourced from SemMedDB~\cite{kilicoglu2012semmeddb}. 
These pairs were annotated by medical experts with prior experience in medical annotations, which covered 20 distinct relation types.
As the nature of the task in triple classification differs from triple extraction, in our study, we initially selected 4,352 positive data instances from MTCD as the source dataset and relabeled the positive data instance.


\subsection{Non-Drug Therapy Annotation}
The CIHT (Complementary and Integrative Health Triples) dataset focuses on the relationship between complementary and integrative health (CIH) therapies and their impact on diseases, genes, gene products, and chemicals.
We defined CIH entities and relation types based on CIHLex~\cite{zhou2023complementary}. To collect data, we utilized all terms in CIHLex to search articles from the abstracts of the PubMed~\cite{white2020pubmed} bibliographic database related to CIH.
From the initial pool of articles retrieved using CIHLex terms, we narrowed down our selection by refining the retrieved abstracts using PubTator~\cite{wei2013pubtator}. This refinement aimed to identify abstracts containing terms related to diseases, genes, and chemicals. We randomly selected 400 abstracts to be included in our dataset.
Subsequently, experts with a Doctor of Chiropractic (DC) degree annotated the dataset according to our pre-defined guidelines. For further details regarding the guidelines, please refer to Appendix~\ref{Annotation_Guidelines}.
To assess the annotation quality, in line with prior work~\cite{zhou2023complementary}, two annotators independently reviewed a shared set of 10\% of the notes. Each annotator then individually annotated the remaining notes. Subsequently, we utilized Cohen's Kappa for token-based annotation to evaluate the inter-annotator agreement for the CIHT dataset annotations.
The Cohen's Kappa score was determined to be 87.99\%.



\subsection{Data Statistics and Comparison with Other Datasets}
In Table~\ref{con:Comparison of GIT}, we show the statistics for GIT as well as the comparable
values for seven   widely-used biomedical triple extraction  datasets,  BioRelEx~\cite{khachatrian2019biorelex}, ADE~\cite{gurulingappa2012development}, CHEMPROT~\cite{taboureau2010chemprot}, DDI~\cite{segura2013semeval}, COMAGC~\cite{lee2013comagc}, EUADR~\cite{van2012eu},
PolySearch~\cite{cheng2008polysearch}. 
GIT  differs from other triple extraction datasets because it includes a broader array of relation types, encompassing 22 distinct types. Additionally, the GIT dataset consists of 4,691 labeled sentences, surpassing the size of all other datasets. This demonstrates
GIT provides a valuable benchmark for biomedical triple extraction. For detailed statistics on each relation type and the respective definitions, please refer to the Appendix~\ref{Statistics for Each Relation Type}.

\begin{table}[ht]
	\centering
	\renewcommand\arraystretch{1.3}
	\scalebox{0.55}{
	\begin{tabular} {ccccc}
		\hline 
		Dataset& \# Entities &  \#Relation Types&\# sentences \\ 
		\hline	
     BioRelEx~\cite{khachatrian2019biorelex} &9,871  & 1 &  2,010 \\
     ADE~\cite{gurulingappa2012development}&  11,070& 2 &  4,272 \\
      CHEMPROT~\cite{taboureau2010chemprot}&  --&  14 & 3,895   \\
        COMAGC~\cite{lee2013comagc}& 541 &  15 &  821  \\
           EUADR~\cite{van2012eu}& 339& 4  &  355  \\
            PolySearch~\cite{cheng2008polysearch}& 255 & 2  &  522  \\
    \textbf{GIT(our dataset)} &  \textbf{5,644} & \textbf{22}  &  \textbf{4,691}  \\
  
		\hline
	\end{tabular}
 }
		\caption{Comparing GIT to the seven commonly used biomedical triple extraction datasets. GIT contains 3,734 training instances, 465 testing instances, and 492 validation instances. In GIT, the training, testing, and validation datasets each consist of distinct instances, ensuring there are no duplicates or overlaps between them.}
	\label{con:Comparison of GIT}
\end{table}

\section{Experiments}
\label{experiments}

\subsection{Evaluation Metrics}
Same as \cite{tang2022unirel,zeng2019learning}, triple is regarded as correct when its relation type, the
head entity and the tail entity are all correct. For example, in the sentence: \textit{Infusion of prostacyclin (PGI2) reportedly attenuates renal ischemic injury in the dog and the rat.},  triple \textit{<Infusion, treats, rat>} is regarded as correct while \textit{ <injury, treats, rat>} is not. 
Following the evaluation method of the previous work~\cite{tang2022unirel,shang2022onerel,lu2022unified,gao2023easy},  we evaluated all the models and reported the evaluation metric, including Micro Precision, Recall, and F1-score.


\subsection{Datasets}
The constructed biomedical triple extraction dataset GIT is used as the benchmark dataset. To further validate the universality of our framework, we also evaluate our model on two commonly utilized biomedical datasets: CHEMPROT~\cite{taboureau2010chemprot} and DDI~\cite{segura2013semeval}. For additional data statistics, please refer to Appendix~\ref{con:Data_Statistics_triple}.

\subsection{Baselines}

\textbf{1) GPT-3.5/4\footnote{https://platform.openai.com/docs/models/overview}}: GPT-3.5-turbo (P1), GPT-3.5-turbo (P2), GPT-4 (P1), and GPT-4 (P2). For these models, we formulate prompts to guide the GPT models in generating triples for each input sentence, along with providing the corresponding relation definitions in the prompts. The distinction between prompt (P1) and prompt (P2) lies in the output format. For more detailed information, please refer to Appendix~\ref{GPT-instruction}. 
We also include the \textbf{2) LLaMA family} as baselines, namely MedLLaMA 13B~\cite{wu2023pmc} and LLaMA2-13b~\cite{touvron2023llama}.

\subsection{Results and Discussion}
\label{con:discussuion_triple}
Table~\ref{con:con:Model main performance} presents the experiment results of various approaches based on Precision, Recall and F value. Results are the average over 5 runs.
We have the following observations: 
  (1) GPT-3.5/4 exhibits the lowest performance.  The main reason is that
the reported results are in the zero-shot setting due to the unavailability of open resources. 
 (2)Despite MedLLaMA 13B being trained on the biomedical domain, it still exhibits worse performance than LLaMA2 13B.

\begin{table*}[ht]
	\centering
	\renewcommand\arraystretch{1.3}
\resizebox{0.8\textwidth}{!}{%
	\begin{tabular} {l|ccc|ccc|ccc}
		\toprule 
  
		\multicolumn{1}{c}  {}&\multicolumn{3}{c}  {DDI}& \multicolumn{3}{c}  {ChemProt} & \multicolumn{3}{c}  {GIT} \\
		 Approach & Precision &  Recall & F1 &  Precision &  Recall & F1&  Precision &  Recall & F1 \\
   
	      \midrule
                 GPT-4 (P1)  &  7.35    &  10.00  &   8.47   &  25.00   & 28.00  & 26.42    &  9.51&  9.25  & 9.38\\
                 GPT-4 (P2)  &  5.08    &    9.00&  6.50    &  20.79   &37.00   &     26.61& 9.48 & 9.46   &9.47  \\
                   MedLLaMA 13B~\cite{wu2023pmc} &   76.63   &    76.63   &  76.63       &  52.10  & 49.04 & 50.52 &42.60 & 41.51  & 42.05 \\
                 LLaMA2 13B~\cite{touvron2023llama} &   79.61    &  79.61   &  79.61     &   78.58  &  76.41&  77.48  &61.76  & 56.45  & 58.99\\
               
           \bottomrule

	\end{tabular}
 }
\vspace{+2mm}
\caption{Results of various approaches for biomeidcal triple extraction on DDI,  ChemProt and GIT }
\label{con:con:Model main performance}
\vspace{-1mm}
\end{table*}

\section{Conclusion}
In this paper, we compare four LLM's performance on three biomedical triple extraction datasets.
 Additionally, we create a biomedical triple extraction dataset with extensive relation type coverage and expert annotations.


\bibliography{anthology}
\appendix
\section{Appendix}

\subsection{Data Statistics}
\label{con:Data_Statistics_triple}
Table~\ref{con:data_Statistics_DDI_chemprot} shows the data statistics for CHEMPROT, DDI, and GM-CIHT.  

\begin{table}[ht]
	\centering
	\renewcommand\arraystretch{1.3}
	\scalebox{0.55}{
	\begin{tabular} {ccccc}
		\hline 
		Dataset& \# Entities &  \#Relation Types&\# train/test/dev \\ 
		\hline	
      CHEMPROT~\cite{taboureau2010chemprot}& 5,990&  14 & 4,111/3,438/2,411   \\
        DDI~\cite{segura2013semeval}&13,107  & 5  &   5,154/1,376/1,436  \\
    \textbf{GM-CIHT (our dataset)} & 5,644 &22 &  3,734/465/492 \\
  
		\hline
	\end{tabular}
 }
		\caption{Data Statistics for CHEMPROT, DDI, and GM-CIHT. "train/test/dev" denotes the counts of (sentence, triples) pairs within each training, testing, and development dataset split.}
	\label{con:data_Statistics_DDI_chemprot}
\end{table}

\subsection{Annotation Guidelines}
\label{Annotation_Guidelines}
In Table~\ref{con:part of Annotation Guideline}, we have presented a part of the annotation guidelines about the annotation of relation types. These guidelines align with the definition of relation types.

\begin{table}[]
    \resizebox{0.45\textwidth}{!}{%
\begin{tabular}{p{0.4\columnwidth}p{1.2\columnwidth}}
\toprule
\textbf{Relation Type}       & \textbf{Definition}   \\                      \hline                                                                  ASSOCIATED WITH &
  CIH therapies that have a correlation or connection with specific chemicals or genes,  either directly or indirectly, without necessarily altering their function. \\
\hline
DISRUPTS &
  CIH therapies that interfere with or disturb the normal function  or balance of particular chemicals or genes, either intentionally or unintentionally. \\
\hline
INHIBITS       & CIH therapies that suppress or reduce the production, release, or activity of certain chemicals or genes.        \\
\hline
STIMULATES     & CIH therapies that promote or enhance the production, release, or activity of specific chemicals or genes.       \\
\hline
TREATS         & CIH therapies applies a remedy with the diseases or symptoms of effecting a cure or managing a condition.        \\
\hline
DOES NOT TREAT & CIH therapies do not applies a remedy with the diseases or symptoms of effecting a cure or managing a condition. \\
\hline
AFFECTS &
  Refers to the direct or indirect influence or impact (positive or negative) that CIH therapies have on the disease or syndrome, its symptoms, or the overall well-being of the individual.\\
\bottomrule

\end{tabular}
}
\vspace{+2mm}
\caption{A part of annotation guideline.}
\vspace{-5mm}
\label{con:part of Annotation Guideline}
\end{table}

\subsection{Statistics for Each Relation Type and the Definition of Each Relation Type}
\label{Statistics for Each Relation Type}
In Table \ref{con:Statistics of GM-CIH}, we present the sentence statistics for our 22 defined relation types within GM-CIHT.
\begin{table}[ht]
	\centering
	\renewcommand\arraystretch{1.3}
\resizebox{0.5\textwidth}{!}{%
	\begin{tabular} {c|c|c|c}
		\toprule 
		 \textbf{relation type} & \textbf{Sentences} & \textbf{relation type} & \textbf{Sentences}   \\ 
	      \midrule
        INTERACTS WITH&   1019  & PREVENTS  &   107 \\
         TREATS&  726   &  PRECEDES &  103  \\
          PROCESS OF&   686   & COMPLICATES  &  101   \\
           INHIBITS&    345  &  ASSOCIATED WITH &  89 \\
      STIMULATES&  298  &   CAUSES&   76 \\
      USES&   293   &  PREDISPOSES& 61   \\
   COEXISTS WITH&   256   &  MANIFESTATION OF &  54  \\
    ADMINISTERED TO&  175    &  AUGMENTS &   53 \\
     DIAGNOSES&    152  &   DISRUPTS&  51  \\
   AFFECTS&  117   &  DOES NOT TREAT &  24  \\
    PRODUCES&   116  &   SYMPTOM OF&   10 \\  
    \bottomrule
  \hline
	\end{tabular}
 }
\vspace{+2mm}
\caption{Statistics of GM-CIHT.}
\label{con:Statistics of GM-CIH}
\vspace{-5mm}

\end{table}
We adapt relation types used by the SemRep biomedical NLP tool~\cite{kilicoglu2020broad}, which is itself adapted from the UMLS Semantic Network. We list the definitions of 22 relation types below~\cite{kilicoglu2011constructing}: 
\begin{enumerate}
\item  CAUSES: Brings about a condition or an effect. Implied here is that an agent, such as for example, a pharmacologic substance or an organism, has brought about the effect. This includes induces, effects, evokes, and etiology. Neurocysticercosis (NCC) is one of the major causes of  neurological disease
\item COMPLICATES: Causes to become more severe or complex, or results in adverse effects.
\item USES: Employs in the carrying out of some activity. This includes applies, utilizes, employs, and avails.
\item STIMULATES: Increases or facilitates the action or function of (substance interaction).
\item DISRUPTS: Alters or influences an already existing condition, state, or situation. Produces a negative effect on.
\item TREATS: Applies a remedy with the object of effecting a cure or managing a condition.
\item COEXISTS\_WITH: Occurs together with, or jointly. Food intolerance-related constipation is characterized by proctitis.
\item  MANIFESTATION\_OF: That part of a phenomenon which is directly observable or concretely or visibly expressed, or which gives evidence to the underlying process. This includes expression of, display of, and exhibition of.
\item INTERACTS\_WITH: Substance interaction.
\item  ADMINISTERED\_TO: Given to an entity, when no assertion is made that the substance or procedure is being given as treatment.
\item  PREVENTS: Stops, hinders or eliminates an action or condition.
\item  PREDISPOSES: To be a risk to a disorder, pathology,or condition.
\item  INHIBITS: Decreases, limits, or blocks the action or function of (substance interaction).\
\item  AUGMENTS: Expands or stimulates a process.
\item PRODUCES: Brings forth, generates or creates. This includes yields, secretes, emits, biosynthesizes, generates, releases, discharges, and creates.
\item  PROCESS\_OF: Disorder occurs in (higher) organism.
\item PRECEDES: Occurs earlier in time. This includes antedates, comes before, is in advance of, predates, and is prior to.
\item  AFFECTS: Produces a direct effect on. Implied here is the altering or influencing of an existing condition, state, situation, or entity. This includes has a role in, alters, influences, predisposes, catalyzes, stimulates, regulates, depresses, impedes, enhances, contributes to, leads to, and modifies.
\item DIAGNOSES: Distinguishes or identifies the nature or characteristics of.
\item ASSOCIATED\_WITH: Has a relationship to (genedisease relation).
\item  DOES\_NOT\_TREAT: antonyms of TREATS
\item SYMPTOM\_OF: departure from normal function or feeling which is noticed by a patient, reflecting the presence of an unusual state, or of a disease; subjective, observed by the patient, cannot be measured directly

\end{enumerate}

\subsection{Instruction of Triple Extraction  by GPT3.5/4}
\label{GPT-instruction}
\begin{figure}[t]
        \centering
        \includegraphics[width=1.0\columnwidth]{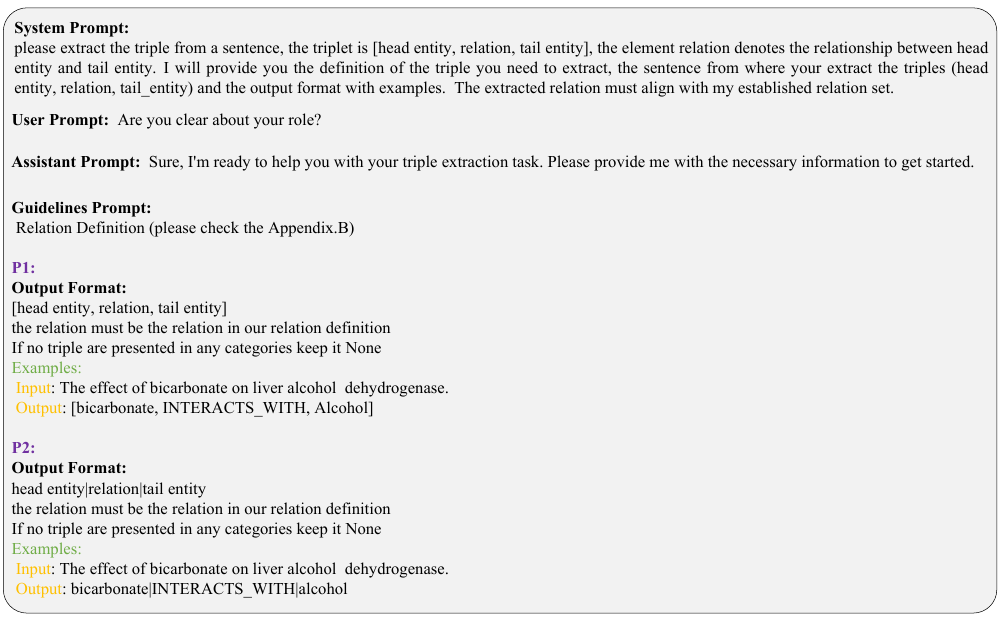}
	\caption{Example of Prompt 1 and Prompt 2 defined for GPT3.5/4  on the task of triple extraction.}
	\label{con:chatgpt_TE}
\end{figure}

Instructions for triple extraction using the GPT API can be found in Figure~\ref{con:chatgpt_TE}.

\end{document}